\documentclass{tADR2e}

\usepackage{subfigure}
\usepackage{enumerate}
\usepackage{color}
\usepackage{cancel}
\usepackage{umoline}
\usepackage{eproofread}
\usepackage{float}
\usepackage{url}
\usepackage{amsmath} 
\usepackage{amssymb}  

\newcommand{\argmax}{\mathop{\rm argmax}\limits}
\newcommand{\V}[1]{\boldsymbol{#1}}
\def\thline{\noalign{\hrule height 1pt}}

\begin{document}



\title{Particle Filter on Episode}

\author{Ryuichi Ueda$^{a}$$^{\ast}$\thanks{$^\ast$Corresponding author. Email: ryuichi.ueda@p.chibakoudai.jp. This work is supported by JSPS KAKENHI Grant Number JP17K00313.},
Masahiro Kato$^{a}$, and Atsushi Saito$^{a}$\\
$^{a}${\em{Chiba Institute of Technology, 2-17-1 Tsudanuma, Narashino, Chiba, Japan}}
}

\maketitle

\begin{abstract}
	Differently from animals, robots can record
	its experience correctly for long time. We propose a novel
	algorithm that runs a particle filter on the
	time sequence of the experience. It can be applied to
	some teach-and-replay tasks.
	In a task, the trainer controls a robot, and the robot
	records its sensor readings and its actions.
	We name the sequence of the record an episode,
	which is derived from the episodic memory of animals.
	After that, the robot executes the particle filter
	so as to find a similar situation with the current one
	from the episode. If the robot chooses the action taken
	in the similar situation, it can replay the taught behavior.
	We name this algorithm the particle filter on episode
	(PFoE). The robot with PFoE shows not only a simple
	replay of a behavior but also recovery motion
	from skids and interruption.
	In this paper, we evaluate the properties of PFoE
	with a small mobile robot.
\end{abstract}

\section{Introduction}\label{sec:introduction}

Animals are receiving vast amount of information from their sense organs.
Their brains may discard most of the information. 
However, they can convert a part of information into abstract knowledge. 
For example, researchers in neuroscience have found that
some special types of neurons in a mammal compose maps of its environment
\cite{okeefe1971,buzsaki2013}.
With the maps, mammals recognize where they are and
where/how they should go.
Even if they forget their past experiences,
they can recognize places.

Mobile robots can also build and utilize maps of their environments
with SLAM (Simultaneous localization and mapping) software\cite{thrun2005}.
In the process of a SLAM algorithm, a robot periodically
receives a set of data from its sensors and
its odometry or dead reckoning module.
A SLAM algorithm converts this data sequence
into a map. Though the sequential order
gives important information to the SLAM algorithm,
it is not recorded in the map.

Though there are similarities in the process of
map building between animals and robots, 
a difference also exists. 
When a robot is recording a data sequence
on the main memory of its computer,
it never disappears until it is explicitly erased.
When a person must memorize a long sequence of numbers
correctly, he/she feels
difficulty. 
On the other hand, robots can memorize a long data sequence
if they have a large amount of DRAM,
and can recall each data in the sequence correctly. 
Since a raw data sequence for a map contains
more information than the map, 
there is a possibility that
we can devise some decision making algorithms
which directly use the sequence.
To make a robot use such an algorithm in real time, 
we must solve the problem of calculation amount.

This paper reports that a simple particle filter
applied to data on a time sequence has 
the ability of real-time decision making in the real world.
This particle filter is named PFoE (particle filter on episode). 
Particle filters\cite{gordon1993}
have been successfully applied to real-time mobile robot localization
as the name {\it Monte Carlo localization} (MCL)\cite{dellaert1999,fox2003}.
MCLs are mainly applied in the state space of a robot,
while PFoE is applied in the time axis.
This paper shows some experimental results.
The robot performs some teach-and-replay tasks
with PFoE in the experiments.
The tasks are simple and could also be handled
by some deep neural network (DNN) approaches\cite{hirose2018,pierson2017}.
However, unlike DNN, the particle filter generates motions
of robots without any learning phase for function approximation.
Moreover, unlike MCL, it never estimates state variables directly. 
This phenomenon has never been reported.

The structure of this paper is as follows. 
Section \ref{sec:works} describes related works.
Section \ref{sec:problem} gives the problem that PFoE solves.
The algorithm is explained and examined in Sections \ref{sec:pfoe}
and \ref{sec:exp} respectively.
We conclude this paper in Section \ref{sec:conclusion}.

A part of the work in this paper derives
from our conference paper \cite{ueda2018icra},
and Japanese conference papers \cite{ueda2017rsj,kato2017rsj}.
We use a generalized definition in the explanation
of the method in this paper. Moreover, we add experiments for
quantitative evaluations and discussion.

\section{Related works}\label{sec:works}

PFoE uses a particle filter, 
which is frequently used for self-localization
of a robot
\cite{dellaert1999,fox2003}.
Particle filters for self-localization
are known by the name of Monte Carlo
localization (MCL). MCL concentrates its
particles in some areas where 
its owner may exist. The particles
are candidates of the actual poses
of the robot. Since the calculation amount
is in proportion to the number of particles,
MCL can work in a large environment.
PFoE also utilizes this advantage of
particle filters.
Though MCL needs a map of the environment
in which a robot works, PFoE does not need it.

Teach-and-replay for mobile robots is an attractive subject
\cite{chen2006,cherubini2009,krajnik2010,sprunk2013,nitsche2014}.
Its goal is to make a mobile robot
autonomously behave as controlled by a trainer previously.
To make a mobile robot work in an environment
with a conventional way,
we must prepare the map of
the environment with a SLAM (simultaneous
localization and mapping) method \cite{thrun2005,montemerlo2003},
implement a self-localization method such as MCL, 
and write navigation codes. 
If teach-and-replay becomes possible, 
these labor hours disappear.

Teach-and-replay for mobile robots
and autonomous vehicles has been
studied in the context of vision-based
feedback control
\cite{chen2006,cherubini2009,krajnik2010,nitsche2014}. 
In a teaching phase, a sequence of
images are recorded. After that,
some feature points are extracted from the images.
In the replay phase, the feature points
in each image are compared to those of
the current image. A robot is controlled
as the both sets of feature points
are matched. 
In \cite{sprunk2013}, 
a two-dimensional laser scanner is
used instead of a camera. 
In this study, a scan matching method
is used for feedback control.

In \cite{krajnik2010,nitsche2014}, 
teach-and-replay methods were applied to
long-distance navigation tasks.
These methods divide a path obtained in
a teaching phase into segments.
In a replay phase, the vehicle tries
to move from the start to the end
of each segment by comparison of images.
Especially in \cite{nitsche2014}, 
Nitsche {\it et al.} have used MCL for
measuring the rate of progress in a segment.
This MCL does not use a global coordinate system,
and estimates the rate of progress with
dead reckoning.
There are some points of similarity
between this method in \cite{nitsche2014} and PFoE.
However, PFoE does not use the concept of self-localization,
or absolute/relative positions in a coordinate system.
Though PFoE has never been applied to
long distance navigation, PFoE can generate
various behaviors of a robot in spite of
its primitive structure.
Moreover, PFoE does not need any
explicit implementation of feedback control.
This is a major difference between
PFoE and the others.

PFoE handles a decision making problem
as a hidden Markov model (HMM).
In an HMM \cite{baum1966,bishop2006}, a finite number
of states are defined, and state transitions 
of the system are hidden from an observer.
Instead, the observer can receive
a sequence of observations.
Many algorithms on HMMs have been proposed
for pattern recognition problems\cite{bishop2006}.
Techniques for HMMs are also utilized for
motion recognition or motion generation of robots
\cite{sugiura2011,dawood2014}. 
They have focused their attention on 
recognition of sequences of motion. 
To make a robot move in the real world
with the recognition result, 
feedback control as researched in
the above teach-and-replay methods
will be required. 
In other research
fields, combinations of an HMM model and a particle filter
can be seen \cite{baxter2010,mushtaq2012}.
Their purposes are also recognition.

In \cite{ueda2016ias}, we have proposed another PFoE as an algorithm for reinforcement learning. Possibly we will unify this PFoE and the PFoE in this paper in future. 
However, they are different with each other in this stage.

Differently from recurrent neural network (RNN)\cite{mayer2006},
long short-term memory (LSTM)\cite{hochreiter1997,graves2005},
and other neural network architectures,
PFoE does not use back-propagation learning.
This is a difference between PFoE and them. 
More than that, it is important that a particle filter
has an ability of teach-and-replay.
Though it can only be applied to some easy tasks at this time,
the phenomena caused by PFoE should be reported
as a new discovery. 
Combinations of the idea of PFoE and some state-of-art
methods will be interesting topics in future.

\section{Problem Definition}\label{sec:problem}

We want to teach an autonomous robot various motions or simple tasks
without changing the software and its parameters on the robot.
The robot is taught its motion by a trainer through a wireless game controller.
Here we formulate this problem. 

\subsection{The system}

We assume a time-invariant system with a state space $\mathcal{X}$.
In the system, a sequence of discrete time steps are defined as
$\mathcal{T} = \{\dots, t-2, t-1, t, t+1, t+2\dots \}$.
The state of the system at $t$ is then written as $x_t \in \mathcal{X}$.
There is a robot in this system. This robot does not
have the knowledge on the system except that it is
time-invariant.

The robot can periodically observe its surroundings
and take an action. Here we define the symbols $z_t$
and $a_t$ which denote an observation at $t$ and an action at $t$ respectively.  
Though an \textit{observation} and an \textit{action} are abstract words,
they correspond to a set of sensor readings and a move of actuators
in the experiment section.
Just after the robot finishes the action $a_t$,
it obtains the observation $z_t$.
After that, it chooses the next action $a_{t+1}$. 
We define an \textit{event} 
$e_t = (a_t,z_t)$.

A state and an observation are related with a probability $P(z_t | x_t)$.
The state transition rule also exists as a probability
$P(x_t | x_{t-1},a_t)$.
However, the robot does not know them.

\subsection{Teaching (obtaining an episode)}

To teach behavior, 
a trainer controls the robot
from an initial state.
We define the start time as $t=0$.
During the control, the robot 
records an event 
$e_t = (a_t,z_t)$
at every time step.
The recording starts from $t=1$.
This procedure is named {\it a teaching phase}. 

After a teaching phase, 
the robot memorizes the following sequence:
\begin{align}
	\mathcal{E}_\text{teach} &= \{a_1,z_1,a_2,z_2,a_3,\dots,a_T,z_T\} \nonumber \\
	&= \{e_t | t = 1,2,\dots,T \}, \nonumber
\end{align}
where $T$ is the time step at which the trainer stopped teaching.
We name the sequence $\mathcal{E}_\text{teach}$ the {\it episode}
of this teaching phase.

Though the robot cannot observe the sequence of states
in the teaching, we write it as
\begin{align}
	\mathcal{X}_\text{teach} = \{x_0,x_1,x_2,\dots,x_T\}
\end{align}
for explanation. 
Since we use a mobile robot in the experiments,
a sequence of states is called a path in this paper.


\subsection{Replay (decision making with the episode)}\label{sub:replay}

After the teaching phase, we
want to make 
the robot replay $\mathcal{X}_\text{teach}$
based on $\mathcal{E}_\text{teach}$.
This phase is named {\it a replay phase}.

In this phase, 
the robot obtains an event
$e_\tau = (a_\tau,z_\tau)$
at every time step.
$\tau$ denotes the latest time step,
at which the robot must decide and take
an action $a_{\tau+1}$.

Our target is not a simple replay in which the robot
has only to trace the sequence of actions $a_1,a_2,\dots,a_T$.
The robot must give feedback to its motion
so as to deal with motion errors and sensor noises. 
Then, it must resume the replay just after 
a sudden interruption.

Though it is merely a matter of form, 
we evaluate a path on a replay $\mathcal{X}_\text{replay}$ with
\begin{align}
	J(\mathcal{X}_\text{teach},\mathcal{X}_\text{replay}) \in \Re.
\end{align}
Though we will use various evaluation criteria in Section \ref{sec:exp},
they can be defined with this form.
From the standpoint of the optimal control theory,
an evaluation function should be defined before
the implementation of a controller.
However, it is not fixed since the robot is given various tasks.
Therefore, the optimality of PFoE cannot be discussed
in this paper. Instead, we will show the versatility of PFoE.

\section{Particle Filter on Episode}\label{sec:pfoe}

We propose a particle filter which searches
similar events to the current event $e_\tau$
from an episode $\mathcal{E}_\text{teach}$.
The robot takes an action
by reference to the action(s) at the similar events.
As a result, the robot replays the path on the teaching. 

\begin{figure}[H]
	\begin{center}
		\includegraphics[width=0.7\linewidth]{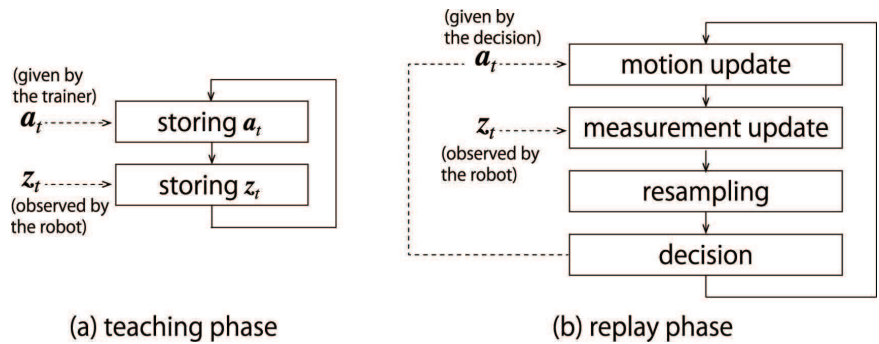}
		\caption{Flow charts of PFoE}
		\label{fig:phases}
	\end{center}
\end{figure}

We show the flow of PFoE in Figure \ref{fig:phases}.
The figure (a) illustrates the teaching phase explained
in Section 3.2. Actions and observations are just
recorded in this phase. In (b), the procedures \textit{motion
update} and \textit{measurement update} are related
to Section 4.2.1 and 4.2.2 respectively.
\textit{Resampling} are briefly written in the end of Section 4.2.2.
\textit{Decision making} is described in Section 4.3.

\subsection{A probability distribution on the episode and its approximation}

We define a probability distribution $Bel_\tau$ which gives the probability
\begin{align}
	Bel_\tau(x_t) &= P(x_\tau \approx x_t | e_\tau, Bel_{\tau-1 })  \nonumber \\
	&= P(x_\tau \approx x_t | a_\tau, z_\tau, Bel_{\tau-1 }) 
	\quad (\forall x_t \in \mathcal{X}_\text{teach}). \label{eq:belief}
\end{align}
$Bel_\tau$ is called the belief on PFoE.
This belief quantifies the similarity
between the current state $x_\tau$
and each $x_t \in \mathcal{X}_\text{teach}$.
We should pay attention to that $x_\tau$ never
becomes equal to $\forall x_t \in \mathcal{X}$
if $\mathcal{X}$ is not discrete. 
Therefore, we did not write as
$x_\tau = x_t$ but wrote as $x_\tau \approx x_t$.
The similarity is indirectly defined by
a likelihood function, which is explained later.
In Equation (\ref{eq:belief}), we have assumed that
the belief should be calculated from the previous
belief and the latest event.
$Bel_\tau$ fulfills $\sum_{t=0}^T Bel_\tau(x_t) = 1$.

\begin{figure}[H]
	\begin{center}
		\includegraphics[width=0.6\linewidth]{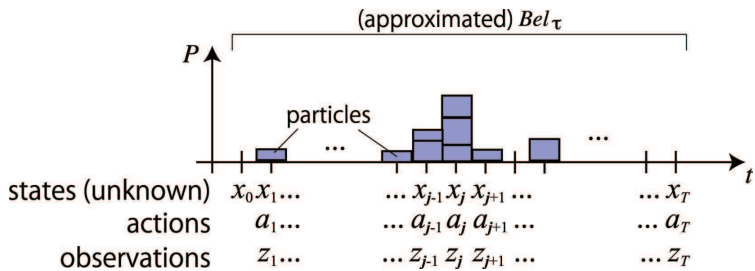}
		\caption{Belief $Bel_\tau$ represented by the particles}
		\label{fig:particles}
	\end{center}
\end{figure}

In PFoE, the belief $Bel_\tau$ 
is approximated by a particle filter which has $N$ particles.
Each of the particles is represented as
\begin{align}
	\xi_\tau^{(i)} = ( t_\tau^{(i)}, w_\tau^{(i)} ) \quad (i=1,2,\dots,N).
\end{align}
In this formulation, $i$ is the index of each particle, 
$t_\tau^{(i)}$ points a time step in the training phase, 
and $w_\tau^{(i)}$ denotes the weight of the particle.
With these particles, each probability $Bel_\tau(x_t)$ is represented by
\begin{align}
	Bel_\tau(x_t) &= \sum_{i=1}^N \delta(x_t = x_{t_\tau^{(i)}}) w_\tau^{(i)} \qquad
	(\delta(\text{true}) = 1 \text{ and } \delta(\text{false}) = 0)\color{blue}.\color{black}
	\label{eq:prob}
\end{align}
Figure \ref{fig:particles} explains this approximation.
The height of each particle represents its weight.
Equation (\ref{eq:prob}) means that the value $Bel_\tau(x_t)$ is equal to
the summation of the heights of the particles at $x_t$.
In this figure, $x_j$ is considered as the most similar state
to $x_\tau$ because $x_j$ has the largest
sum of the weights among the states.

\subsection{State estimation on the episode}

PFoE updates $Bel_{\tau-1}$ after an event
$e_\tau = (a_\tau, z_\tau)$.
At first, the action $a_\tau$ is reflected in the belief. 
This step makes the particles diffuse on the time axis, and
the belief becomes uncertain.
Next, $z_\tau$ is used for reducing the uncertainty. 

\subsubsection{Update with the latest action}

After $a_\tau$, PFoE shifts the belief $Bel_{\tau-1}$
based on $P(x_\tau | x_{\tau-1},a_\tau)$.
This calculation conforms to the following equation of Markov chain:
\begin{align}
	\widehat{Bel}_\tau(x_t) &= P(x_\tau \approx x_t | a_\tau, Bel_{\tau-1}) \nonumber \\
	&= \sum_{t'=0}^{T} P(x_\tau \approx x_t | x_{\tau -1 } \approx x_{t'},a_\tau)Bel_{\tau-1}(x_{t'}), \label{eq:markov}
\end{align}
where $\widehat{Bel}_\tau$ is the belief in which
$z_\tau$ is not reflected.
Since this equation does not take care that the time axis ends at $T$,
it should be cared on the implementation.
The more important problem is that the state transition probability
$P(x_\tau = x_t | x_{\tau -1 } = x_{t'},a_\tau)$ is unknown.
However, when the interval between $\tau-1$ and $\tau$ is short, 
the state transitions will not be largely different from
those of the teaching phase. On the other hand, 
this optimistic assumption should be partially denied.
We prepare a constant number $\Delta$ as the rate of denial
of the assumption.

\begin{figure}[H]
	\begin{center}
		\includegraphics[width=0.6\linewidth]{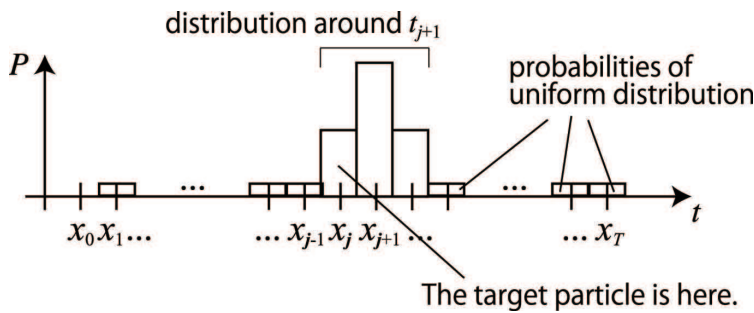}
		\caption{The probability distribution for the update of a particle}
		\label{fig:motion_update}
	\end{center}
\end{figure}

Based on this idea, PFoE updates the belief
from $Bel_{\tau-1}$ to $\widehat{Bel}_\tau$ by
a simple way. 
This procedure changes the time step $t_{\tau-1}^{(i)}$ of each particle
to $\hat{t}_\tau^{(i)}$, which is chosen as
\begin{align}
	\hat{t}_\tau^{(i)} &\sim P(t | t_{\tau-1}^{(i)} )  \label{eq:choose}
	 = (1-\Delta)\cdot P_\alpha(t | t_{\tau-1}^{(i)} ) + \Delta\cdot P_\beta(t). \\
	&\bullet\ P_\alpha: \text{a distribution around } t_{\tau-1}^{(i)} + 1\nonumber \\
	&\bullet\ P_\beta: \text{the uniform distribution on the time axis from $t=1$ to $t=T$}  \nonumber
\end{align}
The first term of the right side of Equation (\ref{eq:choose})
means that the particle at
the time step $t_{\tau-1}^{(i)}$ will go to a time step
near $t_{\tau-1}^{(i)} + 1$
with the probability $1-\Delta$.
With the probability $\Delta$, 
the time step is chosen randomly with $P_\beta$. 
Figure \ref{fig:motion_update} illustrates an example of the probability 
distribution $P(t | t_{\tau-1}^{(i)})$ when
the particle before the update is at $t=t_j$.

Incidentally, when some particles go over the time axis,
they are replaced randomly
on the time axis. Then, any particle is not placed
at $t=0$ because the procedure in Section \ref{sub:obs_update}
cannot handle particles at $t=0$.

\subsubsection{Update with the latest observation}\label{sub:obs_update}

After an observation of $z_t$, the belief
should be updated by the following Bayes' theorem:
\begin{align}
	Bel_\tau(x_t) = P(x_t | z_\tau, \widehat{Bel}_\tau) 
	&= \dfrac{P(z_\tau | x_t)\widehat{Bel}_\tau(x_t)}{\sum_{t'=1}^TP(z_\tau | x_{t'})\widehat{Bel}_\tau(x_{t'})} \nonumber \\
	&= \mu P(z_\tau | x_t)\widehat{Bel}_\tau(x_t), \nonumber
\end{align}
where $\mu$ is the normalizing constant\cite{thrun2005}.
With a likelihood function $L(x_t | z_\tau) \propto P(z_\tau | x_t)$,
the above equation is represented as
\begin{align}
	Bel_\tau(x_t) &= \mu L(x_t | z_\tau)\widehat{Bel}_\tau(x_t). \nonumber
\end{align}
When we use MCL, a likelihood function is
given as a previous knowledge of the robot.
However, there is a problem that
$L(x_t | z_\tau)$ is unknown in the problem defined in Section \ref{sec:problem}.

Instead, PFoE uses $L(z_t | z_\tau)$ as the likelihood function.
Since $z_t$ is the only information directly related to $x_t$
in the episode, this substitution is the only way to
measure the relation between $z_\tau$ and $x_t$.

\begin{figure}[H]
	\begin{center}
		\includegraphics[width=0.6\linewidth]{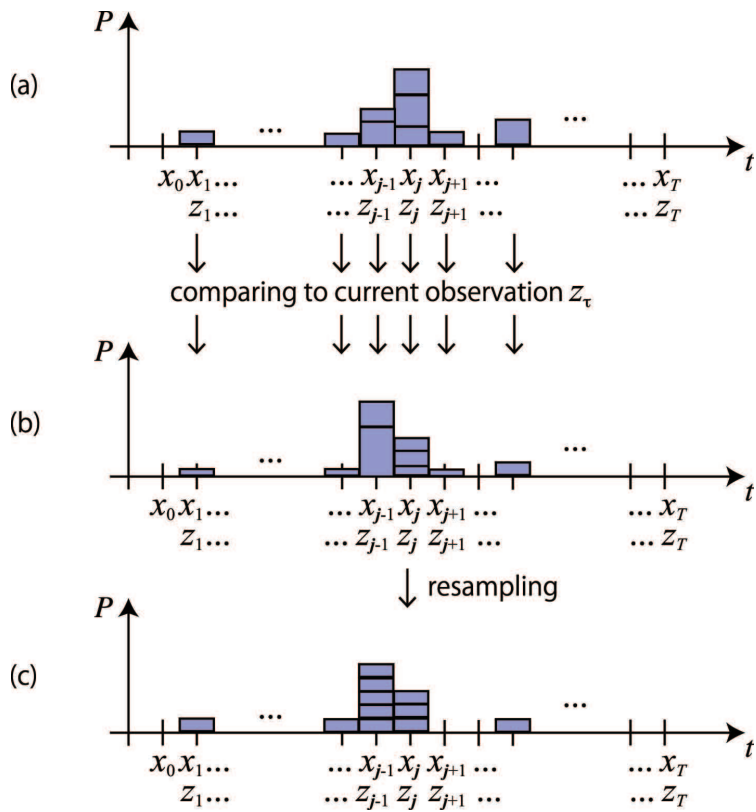}
		\caption{Update after the observation}
		\label{fig:obs_update}
	\end{center}
\end{figure}

With this likelihood function,
PFoE updates the weight of each particle from $w_{\tau-1}^{(i)}$
to $w_{\tau}^{(i)}$ by the following substitution:
\begin{align}
	w_{\tau}^{(i)} = w_{\tau-1}^{(i)}L(z_{t_\tau^{(i)}} | z_\tau). \label{eq:particle_bayes}
\end{align}
This procedure generates the set of particles
${\xi'}_\tau^{(i)} = (\hat{t}_\tau^{(i)}, w_{\tau}^{(i)})\ (i=1,2,\dots,N)$,
which represents $Bel_\tau$.
From Figure \ref{fig:obs_update}(a) to (b),
this procedure is illustrated.

After that, a resampling method
creates another set of particles
$\xi_\tau^{(i)} = (t_\tau^{(i)}, 1/N) \ (i=1,2,\dots,N)$
that also represents $Bel_\tau$.
This procedure eliminates the bias of weights.
It is also used in MCL for the same purpose.
From Figure \ref{fig:obs_update}(b) to (c),
this procedure is illustrated.
We assume that PFoE uses the systematic sampling,
which is usually chosen for MCL\cite{thrun2005}.

\subsection{Decision making}

The next action $a_{\tau+1}$ is chosen based on $Bel_\tau$.
Formally, this decision making is written as 
\begin{align}
	a_{\tau+1} = \Pi(Bel_\tau), \nonumber
\end{align}
where $\Pi$ stands for a mapping from
a probability distribution to an action.
It is called a decision making policy. 

While we cannot fix the one and only implementation of $\Pi$, 
we can formulate some types of policy.
In Section \ref{sec:exp}, we try the following two policies.

The simplest one can be given as
\begin{align}
	\Pi_\text{mode}(Bel_\tau) &= a_{\hat{t} + 1}, \label{eq:mode} \\
	\text{where}\quad \hat{t} &=  \argmax_t Bel_\tau(x_t). \nonumber
\end{align}
Here, $x_{\hat{t}}$ is the mode of $Bel_\tau$.
Therefore, this policy chooses the action
of the next time step of the mode.
We name this policy the {\it mode policy}.

We can also define a policy that uses a mean value of actions
weighted by $Bel_\tau$ if the calculation is possible.
This policy is represented as 
\begin{align}
	\Pi_\text{mean}(Bel_\tau) &= \sum_{t=0}^{T-1} Bel_\tau(x_t) a_{t + 1}. \label{eq:mean} 
\end{align}
When the action can be represented as a vector with some real numbers, 
this calculation is possible.
This policy is named the {\it mean policy}.

\section{Experiments}\label{sec:exp}

We investigate what PFoE can and cannot do
with a mobile robot.
Other methods will not be tried in the
experiments for comparison since they will need
additional experimental assumptions or sensors
as explained in Section \ref{sec:works}.
We will verify that PFoE can work on a small
mobile robot with a small computer and cheap
sensors, and observe various phenomena caused by PFoE.

We have never changed any parameter of PFoE
for all of the experiments in this section.
The setting of the parameters is described
in Section \ref{sub:parameter}.

\subsection{Implementation to an actual robot}\label{sub:parameter}

We implement PFoE to 
Raspberry Pi Mouse\cite{ueda2018ros}, which
is a parallel two-wheel vehicle type
mobile robot.
As shown in Figure~\ref{fig:robot}, 
it has two step motors and four range sensor units.
As the name suggests, it has a Raspberry Pi 3,
which is a well-known single-board computer.
Since we will use this robot as a fully autonomous
one in the later experiments, 
all software programs for this robot including PFoE
work on the Raspberry Pi 3. 
The game controller in the figure is used
for training. 

\subsubsection{Definition of action and observation}

We define an action as
\begin{align}
	\V{a} = (v_\text{linear}, v_\text{angular}), \nonumber
\end{align}
where $v_\text{linear}$ and $v_\text{angular}$
are the linear and angular velocities of the robot respectively. 
This definition is straightforwardly derived from the interface of
a base ROS package (\url{https://github.com/ryuichiueda/raspimouse_ros_2}) of this robot.
This package translates them to the revolution speeds of the step motors.
Since an action becomes a vector in this case, 
we use the bold type $\V{a}$ for representing an action.

In a training phase, we can give $v_\text{linear} = 0.2$[m/s]
to the robot with the up key.
When the up key is not pushed, $v_\text{linear} = 0$.
We can also give $v_\text{angular} = \pm \pi/2$[rad/s]
to it with the right and left keys. 
When these keys are not pushed, $v_\text{angular} = 0$.
These linear and angular velocities
can be given simultaneously.
Since the tires of Raspberry Pi Mouse are very slippy,
these velocities easily decrease by the condition of the floor.

\begin{figure}[H]
	\begin{center}
		\includegraphics[width=0.6\linewidth]{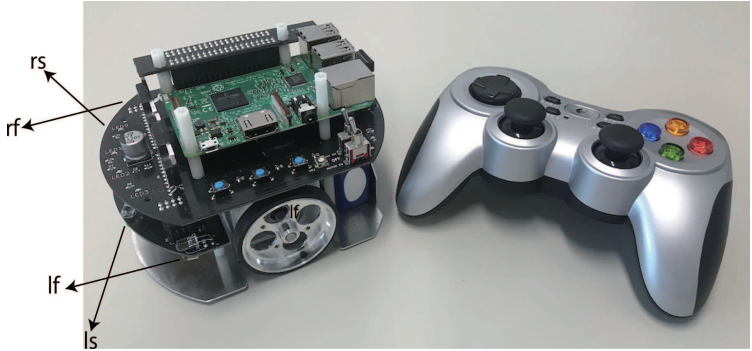}
		\caption{Robot (left) and game controller (right)}
		\label{fig:robot}
	\end{center}
\end{figure}

An observation is defined as 
\begin{align}
	\V{z} = (z_\text{lf},z_\text{ls},z_\text{rs},z_\text{rf}), \nonumber
\end{align}
where the four variables denote the raw values
from the ``left forward (lf),''
``left side (ls),'' ``right side (rs),'' and ``right forward (rf)''
sensors respectively.
Their names are written in Figure \ref{fig:robot}.
Since $\V{z}$ is a vector, we use the bold type.

\begin{figure}[H]
	\begin{center}
		\includegraphics[width=0.6\linewidth]{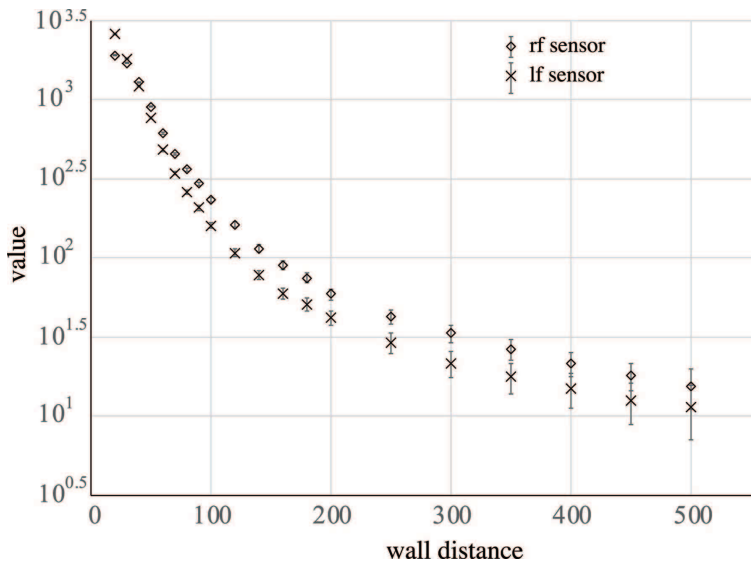}
		\caption{Relation between distances from the wall and sensor values}
		\label{fig:sensor_values}
	\end{center}
\end{figure}

The range sensors are not as accurate as recent
laser range finders.
Each of them is composed of an infrared
LED and a light sensor.
The LED emits infrared light, and the light sensor measures
the intensity of the reflected light.
Since the direction of each sensor can be changed by a touch,
the measured values change easily. 
They are also exposed to vibration when the robot is running.
Moreover, the values are also affected by
the color and material of the wall that reflects the light.

Figure~\ref{fig:sensor_values} shows the character
of the sensors in the case where the robot is not moving.
In the experiment for this data, 
we placed the robot at right
angle to the wall in the corridor used
in Figures \ref{fig:counting} and \ref{fig:square} later.
We changed the distance,
and recorded the values of the lf and rf sensors
every second for one hour with each distance. 
Vertical bars on the data points represent
the standard deviations. 
As shown in the difference between the data of lf and rf, 
each sensor has its own bias even on the logarithmic scale.




\subsubsection{Update procedures}

After an action, each particle 
goes to the next
time step with $30$[\%], moves two steps forward
with $30$[\%], 
does not move
with a probability of $30$[\%], 
or is replaced randomly with $10$[\%].
It means that $\Delta = 0.1$, and that $P_\alpha$
in Equation (\ref{eq:choose}) is the following uniform 
distribution: 
\begin{align}
	 P_\alpha(t | t_{\tau-1}^{(i)} ) =
	 \begin{cases}
		 1/3 & (t = t_{\tau-1}^{(i)}, t_{\tau-1}^{(i)}+1, t_{\tau-1}^{(i)}+2) \\
		 0 & (\text{otherwise})
	 \end{cases}.
\end{align}

After an observation, the weight of each particle is changed with
the following likelihood function:
\begin{align}
	L(\V{z}_t | \V{z}_\tau) 
	= \prod_{j=\text{lf,ls,rs,rf}} \dfrac{1}{|\log_{10}z_{jt} - \log_{10}z_{j\tau}| + 1}, \label{eq:likelihood}
\end{align}
where $z_{jt}$ denotes the value of the $j$ sensor
in $\V{z}_t$.
The value of this function becomes one when $\V{z}_t = \V{z}_\tau$.
We choose this function so as not to
make PFoE sensitive to the difference
when the sensor values are large.


\subsubsection{Other conditions}

Events of the first and last five seconds 
are removed from the episode since
silent events at preparation and finishing
may be contained in these parts.
The number of particles $N$ is fixed to $1000$.
At the start of a replay trial,
particles are randomly placed on 
the episode $\mathcal{E}_\text{teach}$.
The cycle of time steps is fixed to $100$[ms].
The trainer in all of the experiments
is the first author, except that
it is the third author 
in the person following task.
They are accustomed
to controlling the robot. 

\subsection{Counting task}

\begin{figure}[H]
	\begin{center}
		\includegraphics[width=1.0\linewidth]{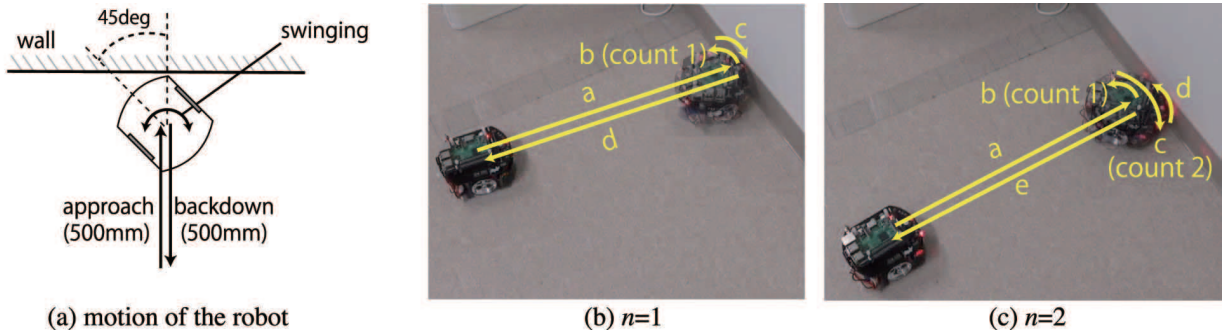}
		\caption{The counting task}
		\label{fig:counting}
	\end{center}
\end{figure}

At first, we make the robot count a number
by its motion. In this task, the trainer
places $500$[mm] apart from the wall.
After that, he makes the robot bump the wall,
swing its nose right and left several times, 
and step back $500$[mm] as illustrated
in Figure \ref{fig:counting}(a).
The trainer repeats this motion several cycles
without a break. 
The angle of the swinging is about $45$[deg]
toward the wall.
We define the number of counts $n$,
which is incremented when the robot's nose starts
leaving the wall by a left or right swing. 
This number is taught to the robot by the control of the trainer.
Figure \ref{fig:counting}(b) and (c) show
the cases where $n=1$ and $n=2$ respectively.

We use this task for evaluating how long PFoE
can trace an episode.
If the robot miscounts in a cycle of replay,
we consider this cycle as a failure. 
The larger the value of $n$, 
the longer the robot must replay
the sequence of actions. 
Miscounts will occur when the mode of the particles
leaps from an event to another due to the
random resampling, sensor noises,
or motion errors.
Moreover, the robot must approach the wall and back down
from it with a right angle. If the robot crossly
approaches the wall, the trace of the episode
falls into disorder. Obviously, 
the robot must decide its action
in consideration of the temporal context of
this task. It is impossible by a decision making
method that decides the robot's action
only based on the latest sensor values. 

\subsubsection{Result}

For evaluation, we had five sets of teach-and-replay
for each number of $n$ ($n=1,2,\dots,8$).
In one set, we had three cycles of the motion at teaching,
and conducted ten cycles of replay just after that.
Each cycle at the replay is called a trial hereafter.
In each trial, we recorded whether
the robot counted $n$ correctly or not.

Table \ref{table:counting} shows the number
of successful trials of each set.
This result was obtained with the mode policy. 
The mean policy did not work
well. It sometimes stopped the robot
because Equation (\ref{eq:mean}) offset
the forward motion and the backward motion.

As shown in this table, the robot could count $n$
without any mistake when $n \le 2$. 
Even when $n=6$, it kept a success rate of more than 80[\%].
When $n=7$ and $8$, the percentages were at around half.
In this table, we can see that the success counts of three sets 
are peculiarly smaller than the others on the same rows. 
It seems that their teaching phases had some problems. 
Practically, since we can discard bad teaching results,
we can expect larger success rates than those on the table.

\begin{table}[thbp]
	\tbl{Numbers of success trials on the counting task (mode policy)}
        {\begin{tabular}{l|rrrrr|r}
        \thline
		$n$ & set 1  & set 2 & set 3 & set 4 & set 5 & all 50 replays\\
        \hline
		$1$ & 10  & 10 & 10 & 10 & 10 & 50 (100[\%])\\
		$2$ & 10  & 10 & 10 & 10 & 10 & 50 (100[\%])\\
		$3$ & {\bf 6}  & 9 & 9 & 10 & 10 & 44 (88[\%])\\
		$4$ & 9  & 9 & 10 & 10 & 10 & 48  (96[\%])\\
		$5$ & 9  & 9 & {\bf 4} & 9 & 10 & 41 (82[\%])\\
		$6$ & 8  & 10 & 9 & 7 & 7 & 41  (82[\%])\\
		$7$ & 7  & 6 & {\bf 0} & 8 & 5 & 26 (52[\%])\\
		$8$ & 3  & 6 & 5 & 3 & 7 & 24 (48[\%])\\
        \thline
  \end{tabular}}
\label{table:counting}
\end{table}

\begin{figure}[H]
	\begin{center}
		\includegraphics[width=0.7\linewidth]{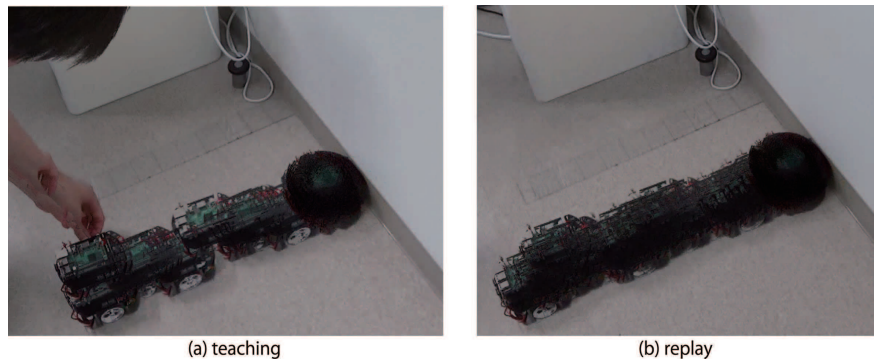}
		\caption{Poses of the robot at the teaching and replay phases}
		\label{fig:n_8_set1}
	\end{center}
\end{figure}

\subsubsection{Behavior of particles}

\begin{figure}[H]
	\begin{center}
		\includegraphics[width=1.0\linewidth]{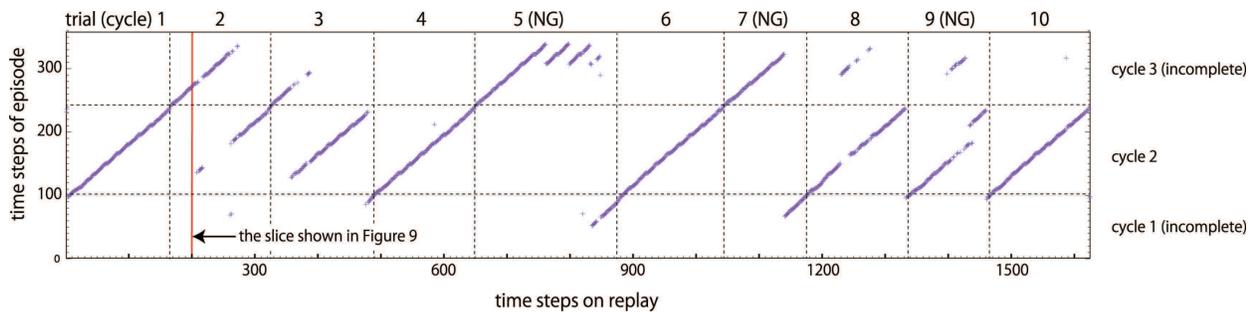}
		\caption{Transition of mode ($n=6$)}
		\label{fig:counting_trace}
	\end{center}
\end{figure}

\begin{figure}[H]
	\begin{center}
		\includegraphics[width=0.5\linewidth]{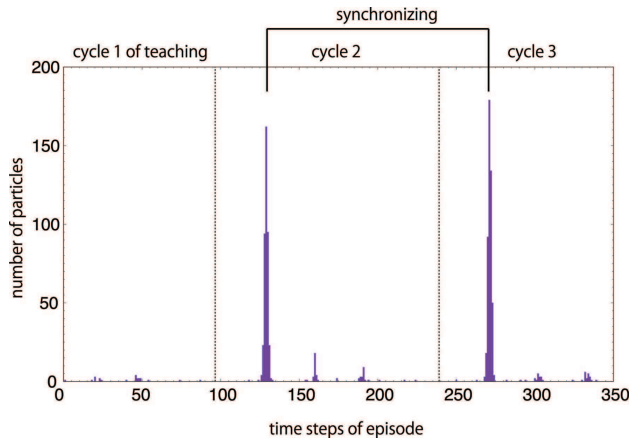}
		\caption{Distribution of particles at a time step on replay}
		\label{fig:distribution}
	\end{center}
\end{figure}

To investigate the behavior of particles in this task, 
we had another set of the teach-and-replay with $n=6$.
Figure \ref{fig:counting_trace} shows
where the mode of the belief existed
at each time step of the trials. 
The horizontal and vertical coordinate values of each point
are time steps at replay and teaching respectively. 
The horizontal and vertical dashed lines
in the figure delimit the cycles (trials). 
We noticed that this episode contained 
only one complete cycle 
due to the five second cut of beginning and end. 
The first and last cycles in this episode
started and ended in the process of counting
respectively.
However, the robot could count the number
successfully seven times in the ten trials.
Figure \ref{fig:distribution} shows 
the distribution of particles at the 200th step
of the replay.
We can find that this distribution
is multi-modal. It has several modes.
The highest one is the mode
shown in Figure \ref{fig:counting_trace}.

As shown in Figure \ref{fig:counting_trace},
the mode of particles leaped frequently.
This leap occurs when the highest mode
in the distribution switches
after an observation. 
When the highest mode changes to
another at the same phase of another cycle,
a mistake does not occur.
However, a miscount happens when
the phases are different.
At the $200$th step, for example, 
the mode existed in the third cycle
of the episode. 
After that, the mode arrived at
the end of episode and disappeared. 
Instead, the second mode in
the second cycle became the main mode.
Since these modes pointed the same phase
of counting, the robot could count correctly 
in this replay cycle. 
In the fifth, seventh, and ninth cycles,
miscounts occurred since 
the phases were not synchronized.
For this problem, 
loop closing techniques used in SLAM
\cite{kawewong2013} will be helpful. 
Then we need to improve our implementation
so as not to cut the start and end of an episode. 
Though these approaches are attractive as future works, 
they may complicate the software.
What is important here is that 
particles in PFoE can trace several points
in an episode, and change the mode
with a degree of accuracy.

\subsection{Choice task}\label{sub:choice}

\begin{figure}[h]
	\begin{center}
		\includegraphics[width=0.4\linewidth]{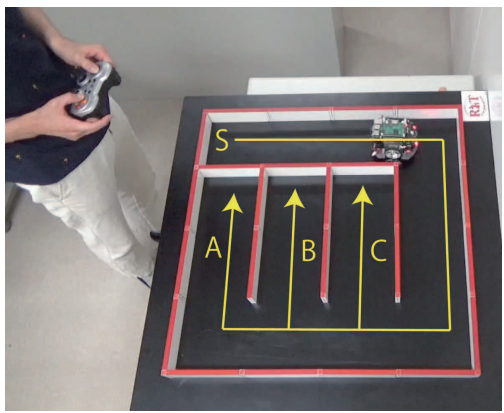}
		\caption{The environment for the choice task}
		\label{fig:choice_task}
	\end{center}
\end{figure}

We have another task in which the robot
must choose a path in a micromouse maze.
The environment is shown in Figure \ref{fig:choice_task}.
In the teaching phase, the trainer makes
the robot run from the start point (``S'' in the figure) to
the end of a target pocket three times.
The target pocket is chosen from A, B, or C in the figure,
and fixed in the three runs.
When the robot reaches the end in the first and second runs,
the trainer replaces the robot at the start
point. Recording of the episode is not stopped
during the replacements.

In the replay phase, we have ten trials.
In a trial, the robot is placed at the start
without any signal. It must start running
autonomously, and must reach the target pocket.
When the center of the robot enters
in the target pocket, 
the trial is regarded as a successful one.
The robot does not need to reach the end deeply.
When the robot stops several seconds 
without entering a pocket, the trainer replaces
the robot at the start point,
and this trial is regarded as a did not finish
(DNF) trial.

We had three sets of teach-and-replay
with the mean policy for each target pocket (nine sets in total).
Then, we had other nine sets with the mode policy.
Incidentally, the trainer had two hours of practice
before this experiment.

\begin{figure}[h]
	\centering
		\includegraphics[width=0.8\linewidth]{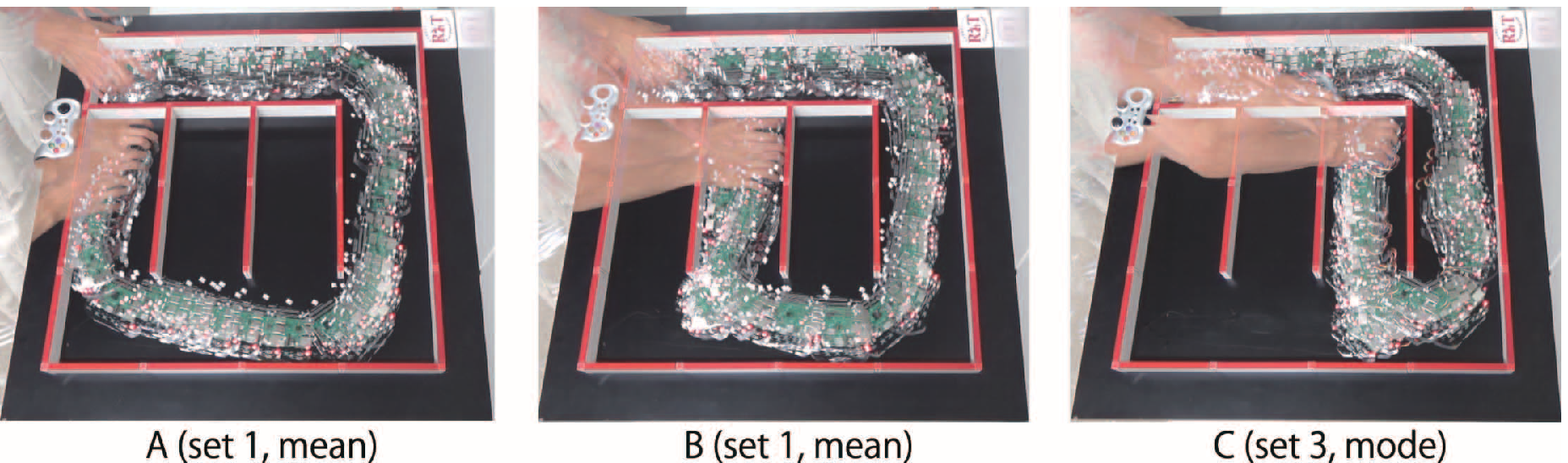}
		\caption{Paths of the robot on the best experimental set for each pocket}
		\label{fig:choice_success}
\ \\
		\includegraphics[width=0.8\linewidth]{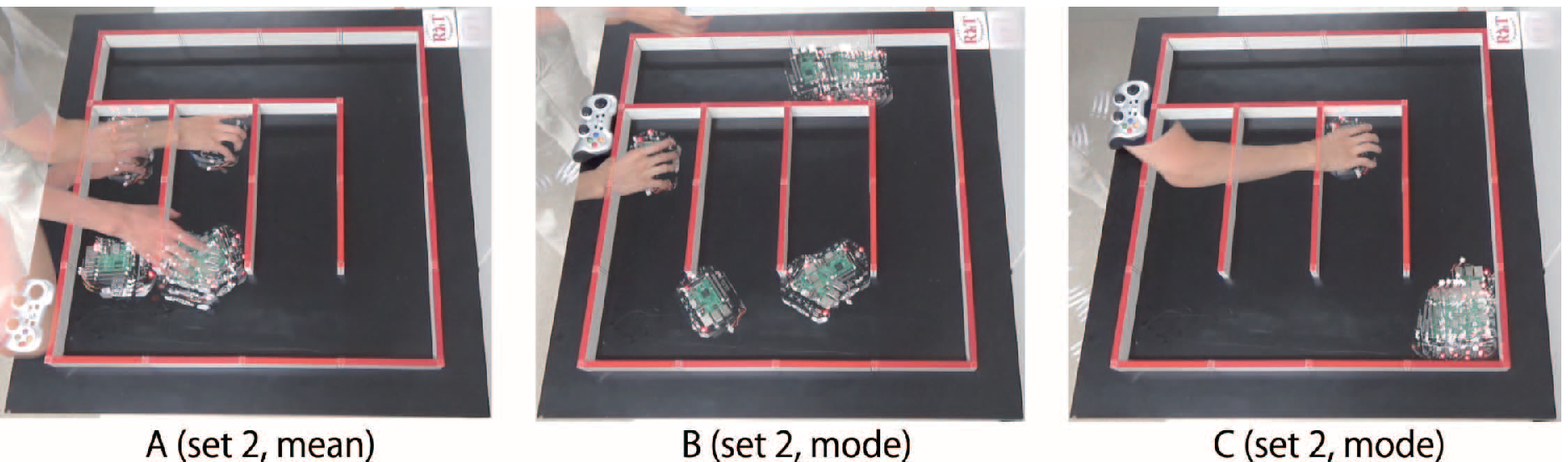}
		\caption{Final poses of the robot on the worst experimental set for each pocket}
		\label{fig:choice_worst}
\end{figure}

The results are illustrated in Table \ref{table:choice}.
The success rates were $83$[\%] and $61$[\%]
with the mean policy and the mode one respectively. 
This percentages will change by the skill or concentration
of the trainer. 
In five sets with the mean policy, the robot did not
mistake. Even with the mode policy, 
the success rates were equal to or more than $90$[\%]
in five sets.
If the trainer can teach appropriately,
the robot can distinguish the three kinds of paths.
Figures \ref{fig:choice_success} shows composite pictures
of the robot in three successful sets respectively. 
The paths in each set of ten trials are drawn in each of them.

Figure \ref{fig:choice_worst}
shows composite pictures of three unsuccessful sets.
In these pictures, the final poses of the ten trials are drawn
respectively. As shown in the second and third pictures, 
the mode policy
frequently made the robot stop before reaching a pocket.
The robot mistook as it finished a trial
frequently when it faced the wall.
That is the reason why DNF trials of the mode
policy were more than those of the mean one.
The robot with the mean policy
also stopped in the same situation.
However, it escaped the situation in many cases
since the mean policy did not make the robot
stop completely.
With a slight motion,
the robot got a change of sensor readings,
and it made a trigger of the escape.

Incidentally, the robot could stop its wheels 
when the trainer picked up it, and start
just after the replacement in every trial. 
Though the explicit start and end
of trials were never recorded in the episodes,
the robot distinguished whether
it was in the maze or not based on the sensor readings.

\begin{table}[thbp]
\tbl{Numbers of success trials on the choice task}
        {\begin{tabular}{c|l|ll}
        \thline
	set  & pocket 	& mean & mode\\
        \thline
		set 1	& A & {\bf 10/10} & \ \ 9/10 \\
		& B & {\bf 10/10} & \ \ 3/10 \\
		& C &  \ \ 7/10  & \ \ 9/10 \\
        \hline
		set 2 & A & \ \ 4/10 & \ \ 5/10\\
		& B & {\bf 10/10} & \ \ {\it 0/10}\\
		& C & \ \ 9/10 & \ \ {\it 1/10}\\
        \hline
		set 3 & A & {\bf 10/10} & \ \ 9/10\\
		& B & {\bf 10/10} & \ \ 9/10\\
		& C & \ \ 5/10 & {\bf 10/10}\\
	\hline
		\multicolumn{2}{l|}{success trials} & 75/90 (83[\%]) & 55/90 (61[\%])\\
		\multicolumn{2}{l|}{DNF trials} & \ 7/90 \  (\ 8[\%]) & 31/90 (34[\%])\\
		\multicolumn{2}{l|}{mischoice trials} & \ 8/90 \  (\ 9[\%]) & \ 4/90 \  (\ 4[\%])\\
		\multicolumn{2}{l|}{success trials in finished trials} & 75/83 (90[\%]) & 55/59 (93[\%])\\
        \thline
  \end{tabular}}
\label{table:choice}
\end{table}

\subsection{Wall following task}\label{sub:wall_following}

\begin{figure}[h]
	\begin{center}
		\includegraphics[width=0.6\linewidth]{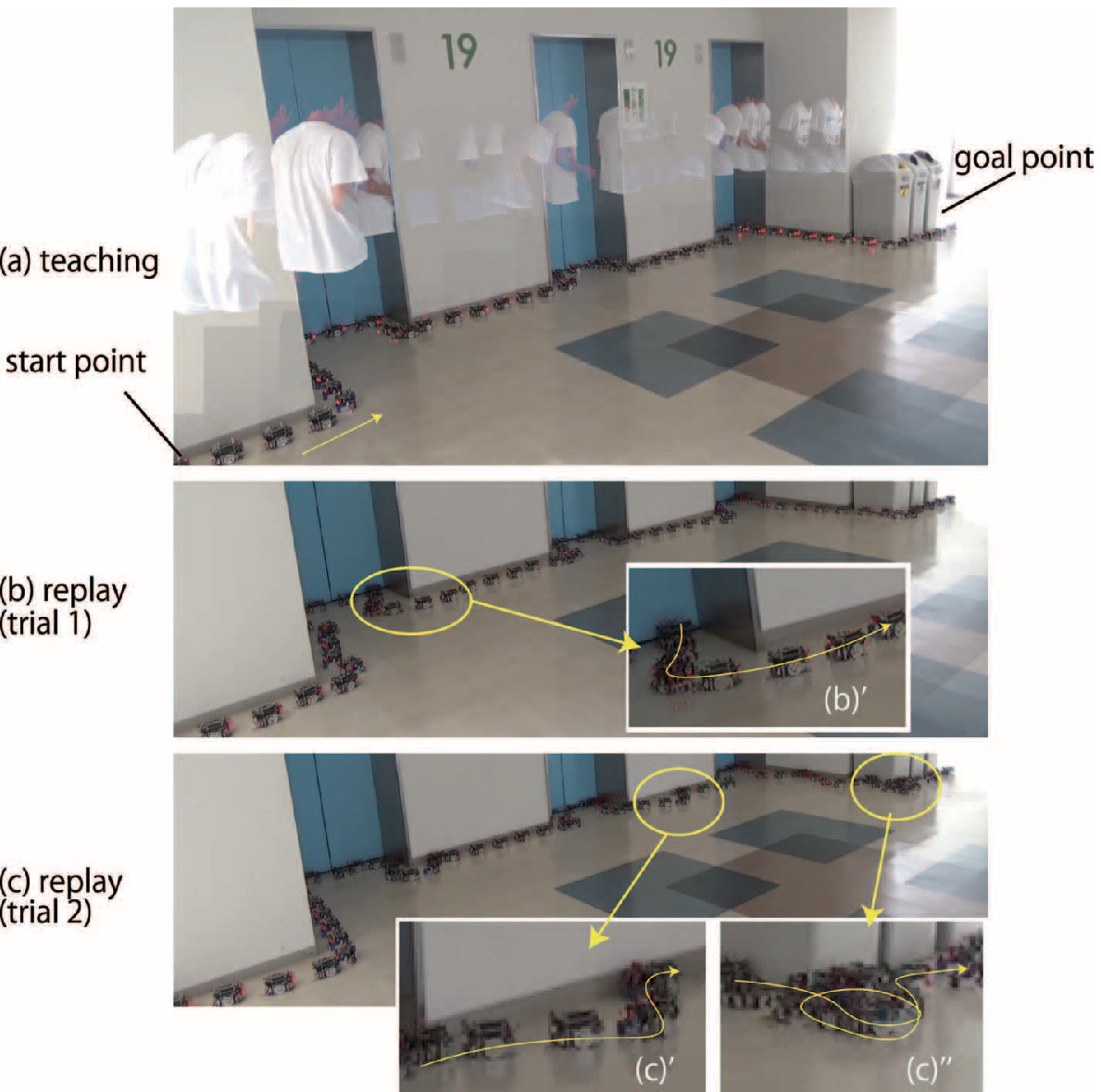}
		\caption{Wall following task}
		\label{fig:wall_trace}
\ \\
		\includegraphics[width=1.0\linewidth]{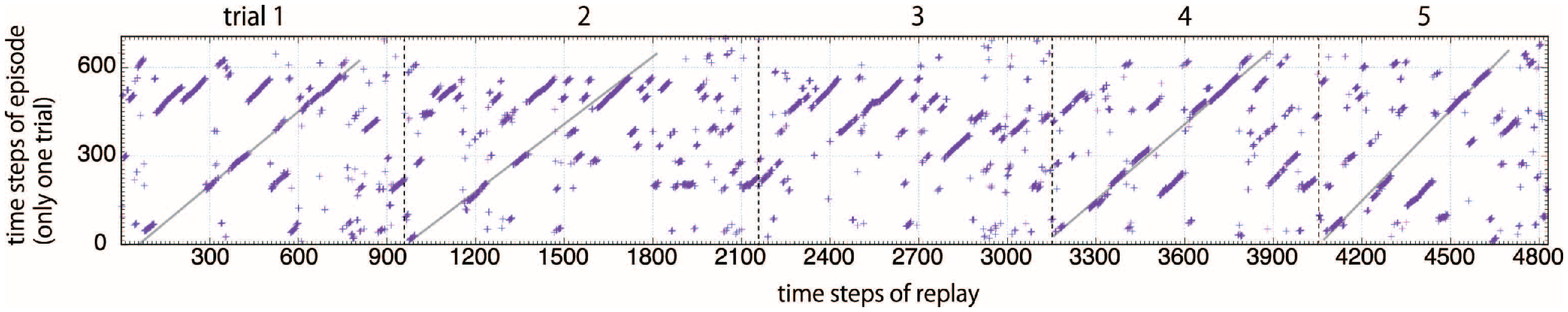}
		\caption{Change of mode in the wall following task}
		\label{fig:wall_trace_mode}
	\end{center}
\end{figure}

Next, we make the robot follow the indented wall
of an elevator hall. In a training phase,
the trainer controls the robot 
as shown in Figure \ref{fig:wall_trace}(a)
only once. At a replay trial,  
we make the robot run from the start
point to the goal point behind the rightmost dustbox.
If the robot can reach the goal point, 
it is regarded as a success.
Though we do not evaluate the success rate
in this task, we aim for five consecutive successful trials
without failure after a training phase. 
We do not give the robot any start signal at the beginning of each trial.

In this task, the robot only has to react 
the latest sensor readings.
Instead, the robot must maintain the follow of the wall
after a skid or a bump against the wall by its feedback motion.
A problem is that 
the ability of PFoE to trace an episode
has the possibility to
inhibit the feedback motion.

We spent about two hours to the experiment.
In the two hours, 
successful sets of teach-and-replay increased progressively
due to the habituation of the trainer.
Though some troubles which were not related
to PFoE occurred in replay trials\footnote{
clogging of log output, actuation of an elevator door,
and halt of a ROS node},  
we obtained one successful teach-and-replay set
with the mean policy and two successful sets
with the mode policy. 

Figures \ref{fig:wall_trace}(b) and (c)
show the first and second replay trials with the episode
obtained by the teaching in Figure \ref{fig:wall_trace}(a).
The mode policy was used for these trials.
In these trials and the other three trials,
the robot successfully followed the wall with its feedback motion. 
In many cases where the robot largely got away from the wall, 
it returned to the wall with behaviors shown 
in the enlarged images (b)' and (c)'.
Though these kinds of large recovery motion
were not directly taught in the teaching,
they were generated by PFoE. 
Figure \ref{fig:wall_trace}(c)'' is an extreme case, 
where the robot ran around in circles
at the corner after it lost the wall
(to be more exact, the dustbox). 
Such a motion was never taught to the robot.

Figure \ref{fig:wall_trace_mode} shows the transition
of the mode in the five trials. 
As shown in this diagram, the mode leaped frequently. 
If these leaps were prohibited, the robot would not
take appropriate feedback.
Though tracing of an episode is the main feature of PFoE,
it can also discard a trace quickly. 

On the other hand, we can find in Figure \ref{fig:wall_trace_mode}
that the mode was synchronizing with the sequence of the training.
Particles of the modes on the gray diagonal lines
have the information of the progress of this task. 
Though this information was not required for this task, 
we could found that PFoE partially sustained the trace of the episode.

\subsection{Wall following and corridor crossing task}\label{sub:problem}

\begin{figure}[H]
	\begin{center}
		\includegraphics[width=1.0\linewidth]{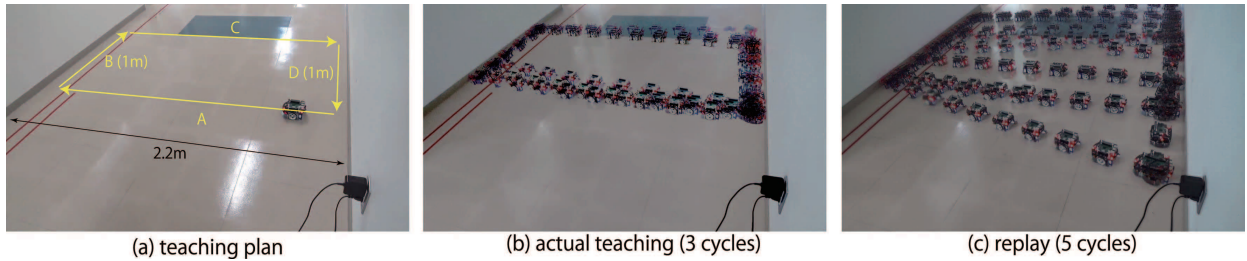}
		\caption{Teaching and replay phases on the wall following and corridor crossing task}
		\label{fig:square}
	\end{center}
\end{figure}

Lastly, we prepare a task that requires
not only the feedback motion as shown in the wall following
but also the trace of an episode. 
In the task named
the wall following and corridor crossing task, 
the robot runs on a rectangle path as shown in
Figure \ref{fig:square}(a). 
During the intervals B and D shown in (a), 
the trainer controls the robot as it follows the wall.
The trainer makes the robot turn 90[deg]
in the ends of B and D, and makes it go straight
in A and C.

After three cycles of a teaching phase, 
we made the robot replay five cycles (trials)
continuously. 
All of the cycles in the teaching and replay
are shown in Figures \ref{fig:square}(b) and (c)
respectively. As shown in (c), the paths
at the replay were cluttered since the lengths
of the wall following varied. 

In Table \ref{table:time_variety},
we wrote out the amounts of time
consumed in the intervals B and D.
We can find that four values in the
row of replay are smaller than the minimum
value in the row of teaching.

\begin{table}[thbp]
	\tbl{Time for intervals B and D}
        {\begin{tabular}{l|cc|cc|cc|cc|cc|c}
	\thline
		$n$ 
		& \multicolumn{2}{c|}{cycle 1} 
		& \multicolumn{2}{c|}{cycle 2} 
		& \multicolumn{2}{c|}{cycle 3} 
		& \multicolumn{2}{c|}{cycle 4} 
		& \multicolumn{2}{c|}{cycle 5} 
		& \\
		interval 
		& B & D 
		& B & D 
		& B & D 
		& B & D 
		& B & D 
		& avg. (std. dev.)\\
        \hline
		teaching & 5.0 & 5.3 & 4.3 & 4.8 & 3.4 & 5.3 &-&-&-&-& 4.7 (0.7)\\
		replay & {\it 1.2} &4.5 &4.4 &5.3 &5.7 & {\it 3.2} & {\it 2.1} &{\it 3.0} &4.2 &3.7 & 3.7 (1.6)\\
        \thline
  \end{tabular}}
\label{table:time_variety}
\end{table}

\begin{figure}[H]
	\begin{center}
		\includegraphics[width=0.7\linewidth]{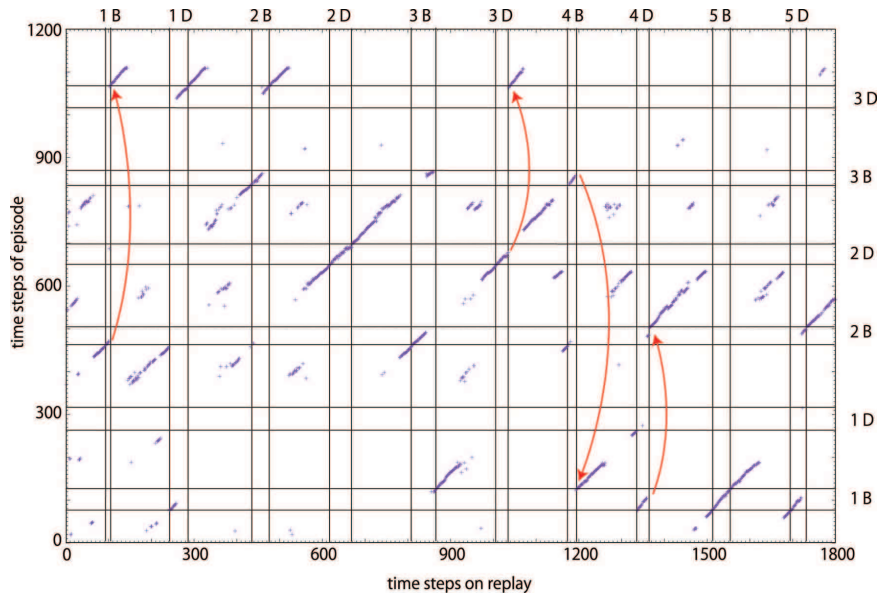}
		\caption{Transition of mode in the set of the wall following and corridor crossing task}
		\label{fig:corridor_rotating}
	\end{center}
\end{figure}

We can understand the reason with Figure
\ref{fig:corridor_rotating}. 
In the intervals that were finished in short time,
the modes leaped too early to beginning time steps of
the intervals A or C.
Since leaps frequently occur in the wall following part
as mentioned in Section \ref{sub:wall_following},  
the information of the time is easily lost.
This is a major problem of the current version of PFoE. 
Though we have already tried an idea for
this problem in \cite{saito2018},
it has never been matured.

Though we also tried the mean policy for this task,
the motion of the robot was inaccurate. 
In this case, the angles when the robot left
from a wall did not become 90[deg].

\subsection{Discussion about the choice of policies}

Equation (\ref{eq:mean}) of the mean policy prevents
the robot from moving
when some clusters of particles
suggest contradictory actions.
This effect was shown in the counting task.
Moreover, Equation (\ref{eq:mean}) obscures
the amount of displacement even when 
the robot must move accurately.
This effect was observed in the wall following
and corridor crossing task.

On the other hand, the mode policy is not the only option.
It just utilizes one of feature amounts of $Bel$. 
We should consider a combination of several statistics of $Bel$
in future works.

\subsection{Other behaviors}

\begin{figure}[H]
	\begin{minipage}{20em}
	\begin{center}
		\includegraphics[width=1.0\linewidth]{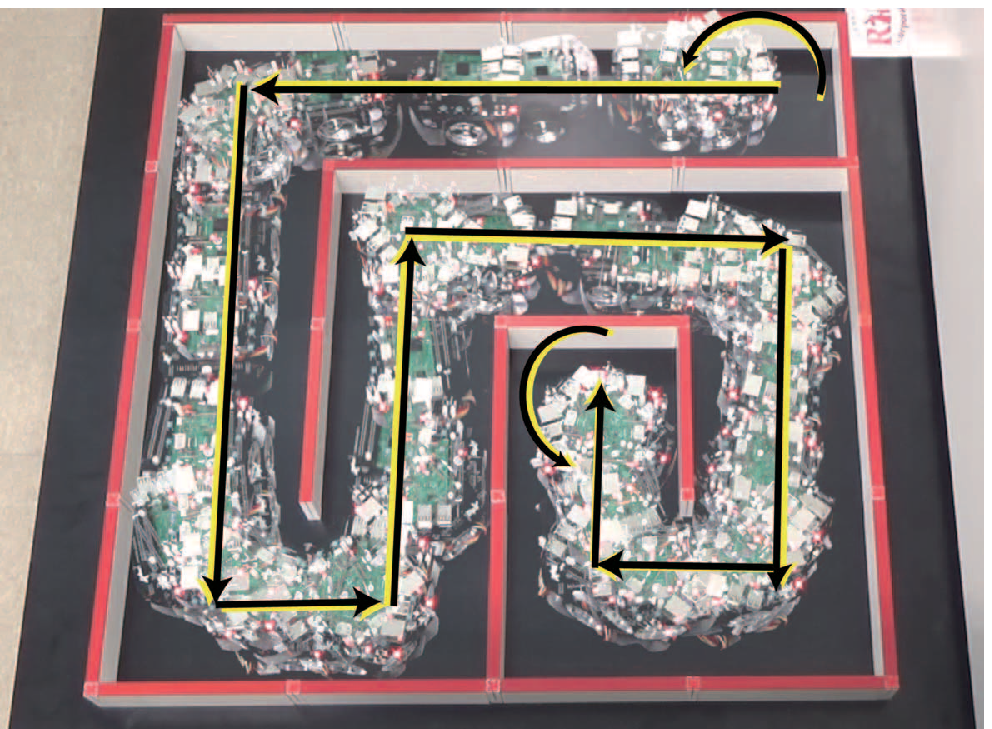}
		\caption{Running in a maze}
		\label{fig:maze}
	\end{center}
	\end{minipage}
	\begin{minipage}{20em}
	\begin{center}
		\includegraphics[width=1.0\linewidth]{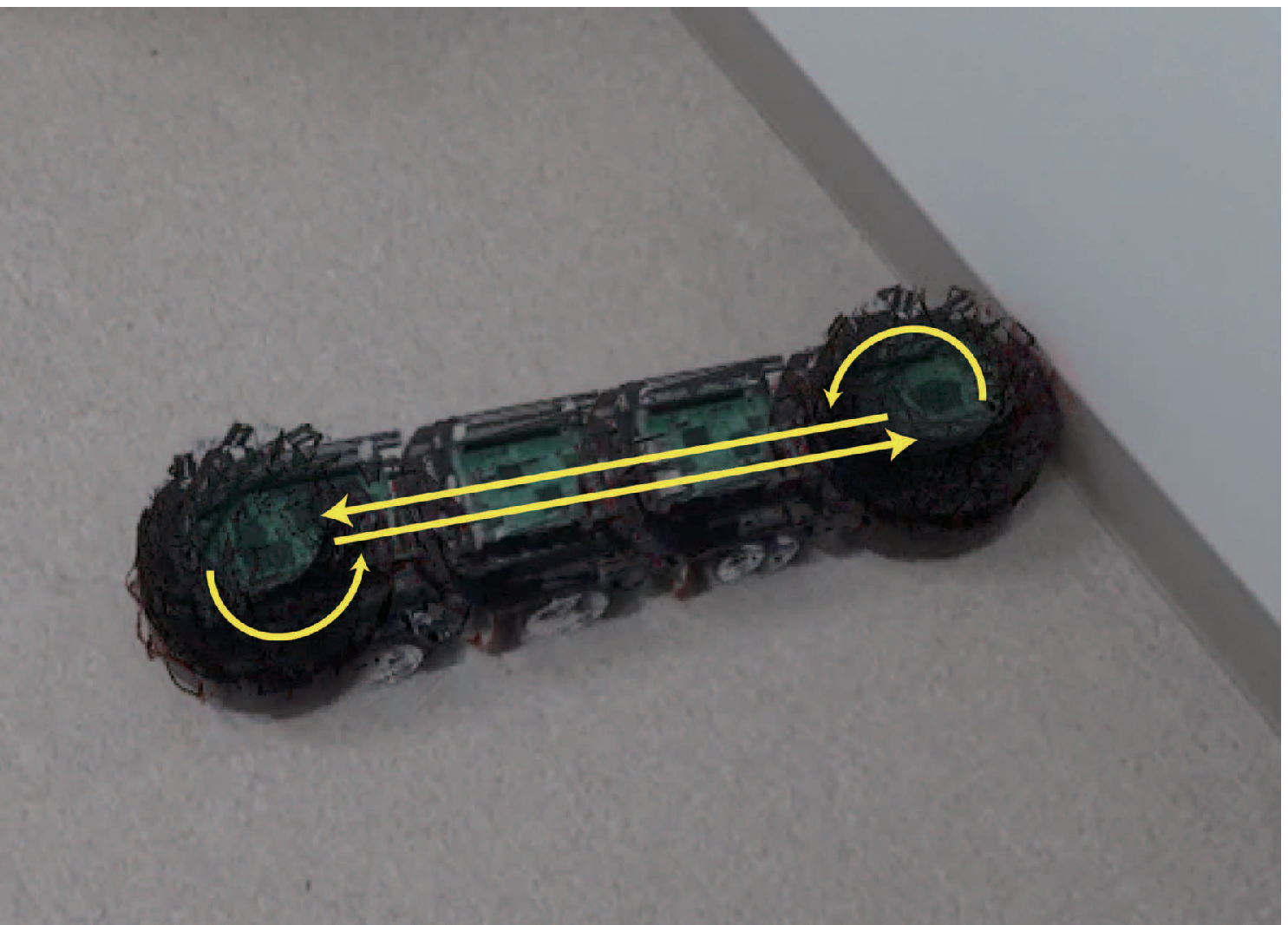}
		\caption{A behavior in front of the wall}
		\label{fig:turn}
	\end{center}
	\end{minipage}
\end{figure}

We have generated various behaviors of the robot with the same parameter set.
Figure \ref{fig:maze} illustrates a cycle of replay in a micromouse maze. 
Since the robot easily skids and hits the walls, 
feedback control is important for stable run. 
When we took the movie for this figure,
we had three cycles of teaching and ten minute replay.
We used the mode policy. 
In the replay, the robot ran the path without stopping,
while the robot took irregular 180-degree turns twice.
Though PFoE should be able to make the robot
trace mazes which have several intersections,
the success rates in the choice task
are not enough for multiple choices.
A high level teaching skill or a patchwork
technique of episodes will be required.

In Figure \ref{fig:turn}, the robot had a
180-turn both at the front of wall
and the farthest point from the wall. 
This is a modified version of the counting task.

\begin{figure}[H]
	\centering
	\begin{minipage}{18em}
	\centering
		\includegraphics[width=1.0\linewidth]{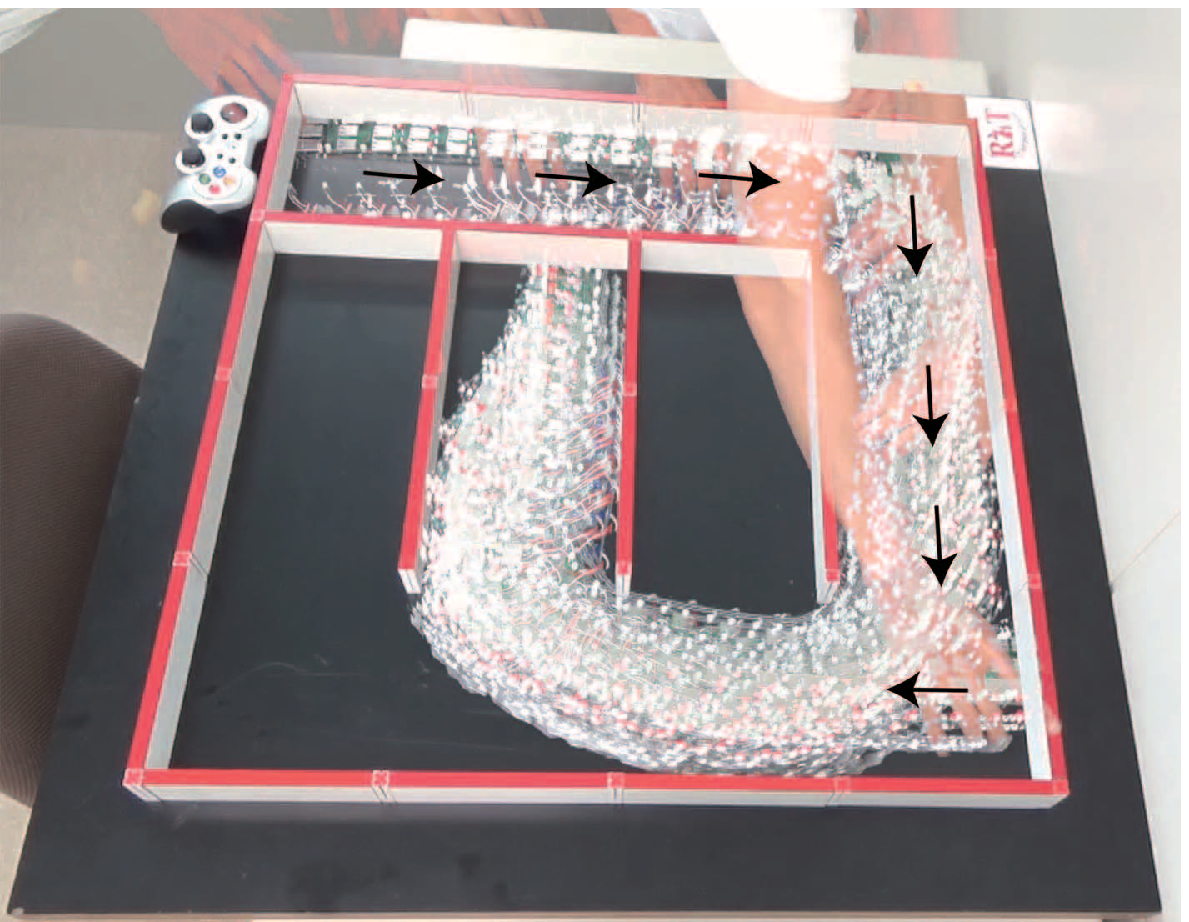}
		\caption{Choice task from different start points}
		\label{fig:change_start}
	\end{minipage}
	\begin{minipage}{18em}
	\centering
		\includegraphics[width=1.0\linewidth]{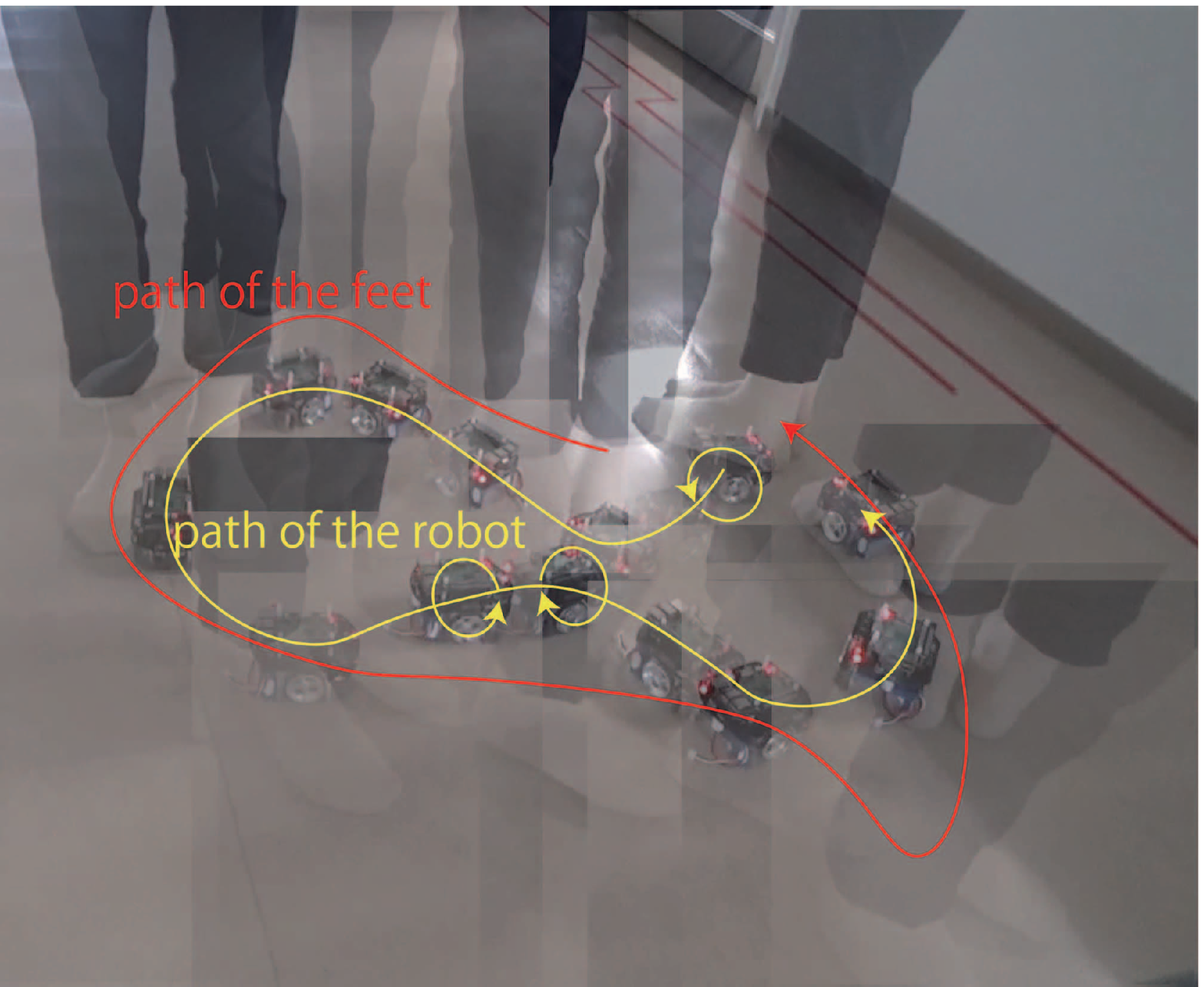}
		\caption{A part of trial of person following}
		\label{fig:person_follow}
	\end{minipage}
\end{figure}

In Figure \ref{fig:change_start}, 
paths of seven consecutive replay trials
are illustrated at the choice task. 
These paths start from the different
points, which are indicated by arrows
in the figure. The training 
was done with the same start point
in Figure \ref{fig:choice_task}.
The target pocket was B. 
The mean policy was used.

As shown in the figure, 
the robot repeatedly chose the target pocket
in spite of the difference of the start points.
This case was the best one, and
the robot sometimes made mistakes in other sets. 
However, this result suggests that PFoE
does not require strict start points
in some tasks. In \cite{ueda2018icra},
we can see another case where
PFoE shows the robustness. 
In this case, the counting task was interrupted 
and restarted from a different point. 
However, the robot could get the rhythm
immediately.

Figure \ref{fig:person_follow} shows
the paths of a person and the robot
in a person following task.
In the teaching phase, the trainer
rotated the robot as a search behavior
when the target person
was far from it (farther than about 30[cm]). 
When the target was near, 
the trainer made the robot follow him.
We spent 80[s] for this training. 
Figure \ref{fig:person_follow} is a composite picture
of some scenes in 80[s] replay.
The three hoops on the path of the robot
mean the points where
the robot took the search behavior.
Though this figure does not tell it,
the robot could appropriately choose
the search behavior and the following behavior
as it taught in the teaching phase. 

\subsection{Calculation amount}

Finally, we measured the time consumption of PFoE
on Raspberry Pi 3, which has a 
ARM Cortex-A53 1.2 GHz processor.
Though this processor has four CPU cores, 
our implementation of PFoE runs on a single thread.

In a measurement, $m$ cycles of the counting task 
were trained. Then we logged the time consumption
of each procedure in PFoE at the replay phase.
Since we did not use a real-time kernel, 
large delays unrelated to the calculation occurred
once in about ten time steps.
We chose five consecutive time steps that did not
contain the delay from the log, 
and took the averages of them. 
We tried $m = 3,10,20$. 

The results are shown in Table \ref{table:time_consuming}.
Since the time complexity of every procedure in PFoE is $O(N)$,
the time length of each procedure never extends by the
increase of the length of the episode.

\begin{table}[thbp]
	\tbl{Time for each step of PFoE on Raspberry Pi 3 (unit: millisecond)}
	{\begin{tabular}{l|ccc}
	\thline
		lengths of teaching & 3 cycles  & 10 cycles & 20 cycles \\
        \hline
		update with an action & 1.51 & 1.56 & 1.55\\
		update with an observation & 4.47 & 4.52 & 4.46 \\
		resampling & 1.27 & 1.29 & 1.28 \\
		decision making (mode policy) & 0.72 & 0.72  & 0.72\\
        \hline
		total & 7.97 & 8.09 & 8.01\\
        \thline
  \end{tabular}}
\label{table:time_consuming}
\end{table}

There is a problem that the calculation amount of
the update with an observation is in proportion
to the number of the sensors. 
As shown in the table, 
the update with an observation
is the heaviest procedure in PFoE
since four sensor values have to be
reflected to the weight of every particle 
one by one. 
Since MCL and online SLAM have exactly the same problem, 
we will be able to utilize the techniques for MCL and online SLAM
\cite{thrun2005}
for it. However, it is still a future work.



\section{Conclusion}\label{sec:conclusion}

We have proposed PFoE,
which uses a particle filter not for mobile robot localization but for
making a robot recall and replay memory of the past. 
Our implementation of PFoE is extremely simple. 
The concepts of states, positions, and distances are not used in it. 
However, it can work on the actual robot. 

In the experiments, we have obtained the following knowledge:
\begin{itemize}
	\item PFoE can make the robot replay behaviors that have temporal
		contexts as shown in the counting task though the length is still limited. 
	\item PFoE can also deal with tasks
		that have spatial contexts as shown in the wall following task. 
	\item PFoE can also make the robot replay behaviors
		that have not only spatial contexts but temporal contexts
		as shown in the counting task.
		In the counting task, the robot could count six successfully
		more than 80[\%] of the time. 
	\item PFoE with the parameters mentioned in Section \ref{sub:parameter} generates
		various motions of the robot. 
	\item A step of PFoE which uses 1,000 particles can be calculated in 8[ms]
		on a 1.2 GHz ARM processor regardless of the length of the episode.
\end{itemize}

We have also confirmed some problems of PFoE through the experiments.
Since the mean and mode policies
are only feature amounts of $Bel$, there is a room for improvement of
policies. Moreover, for applying PFoE to long term tasks, 
unnecessary leaps of the mode should be reduced.



\begin{thebibliography}{10}
\providecommand{\url}[1]{#1}
\csname url@samestyle\endcsname
\providecommand{\newblock}{\relax}
\providecommand{\bibinfo}[2]{#2}
\providecommand{\BIBentrySTDinterwordspacing}{\spaceskip=0pt\relax}
\providecommand{\BIBentryALTinterwordstretchfactor}{4}
\providecommand{\BIBentryALTinterwordspacing}{\spaceskip=\fontdimen2\font plus
\BIBentryALTinterwordstretchfactor\fontdimen3\font minus
  \fontdimen4\font\relax}
\providecommand{\BIBforeignlanguage}[2]{{%
\expandafter\ifx\csname l@#1\endcsname\relax
\typeout{** WARNING: IEEEtran.bst: No hyphenation pattern has been}%
\typeout{** loaded for the language `#1'. Using the pattern for}%
\typeout{** the default language instead.}%
\else
\language=\csname l@#1\endcsname
\fi
#2}}
\providecommand{\BIBdecl}{\relax}
\BIBdecl

\bibitem{okeefe1971}
J.~O'keefe and J.~Dostrovsky, ``{The hippocampus as a spatial map. Preliminary
  evidence from unit activity in the freely-moving rat},'' \emph{Brain
  Research}, vol.~34, no.~1, pp. 171--175, 1971.

\bibitem{buzsaki2013}
G.~Buzs{\'a}ki and E.~I. Moser, ``{Memory, navigation and theta rhythm in the
  hippocampal-entorhinal system},'' \emph{Nature Neuroscience}, vol.~16, no.~2,
  pp. 130--138, 2013.

\bibitem{thrun2005}
S.~Thrun, W.~Burgard, and D.~Fox, \emph{{Probabilistic ROBOTICS}}.\hskip 1em
  plus 0.5em minus 0.4em\relax MIT Press, 2005.

\bibitem{gordon1993}
N.~J. Gordon, D.~J. Salmond, and A.~F.~M. Smith, ``{Novel approach to
  nonlinear/non-Gaussian Bayesian state estimation},'' \emph{IEE
  Proceedings-F}, vol. 140, no.~2, pp. 107--113, 1993.

\bibitem{dellaert1999}
F.~Dellaert, D.~Fox, W.~Burgard, and S.~Thrun, ``{Monte Carlo Localization for
  Mobile Robots},'' in \emph{Proc. of IEEE International Conference on Robotics
  and Automation (ICRA99)}, 1999, pp. 1322--1328.

\bibitem{fox2003}
D.~Fox, ``{Adapting the Sample Size in Particle Filters Through
  KLD-Sampling},'' \emph{International Journal of Robotics Research}, vol.~22,
  no.~12, pp. 985--1003, 2003.

\bibitem{hirose2018}
N.~Hirose, R.~Tajima, and K.~Sukigawa, ``{MPC policy learning using DNN for
  human following control without collision},'' \emph{Advanced Robotics},
  vol.~32, no.~3, pp. 148--159, 2018.

\bibitem{pierson2017}
H.~A. Pierson and M.~S. Gashler, ``{Deep learning in robotics: a review of
  recent research},'' \emph{Advanced Robotics}, vol.~31, no.~16, pp. 821--835,
  2017.

\bibitem{ueda2018icra}
R.~Ueda, M.~Kato, A.~Saito, and R.~Okazaki, ``Teach-and-replay of mobile robot
  with particle filter on episode,'' in \emph{Proc. of IEEE International
  Conference on Robotics and Automation}, Brisbane, Australia, 2018, pp.
  3475--3481.

\bibitem{ueda2017rsj}
R.~Ueda, ``Particle filter on episode for teaching,'' in \emph{{Proc. of the
  35th Annual Conference of the Robotics Society of Japan}}, Kawagoe, Japan,
  2017 (in Japanese).

\bibitem{kato2017rsj}
M.~Kato, R.~Ueda, and R.~Okazaki, ``{An application of Particle Filter on
  Episode for Teaching --- Replay of Path Selection and Motion Control of a
  Mobile Robot by Humans},'' in \emph{{Proc. of the 35th Annual Conference of
  the Robotics Society of Japan}}, Kawagoe, Japan, 2017 (in Japanese).

\bibitem{chen2006}
Z.~Chen and S.~T. Birchfield, ``{Qualitative Vision-Based Mobile Robot
  Navigation},'' in \emph{Proceedings of the 2006 IEEE International Conference
  on Robotics and Automation}, 2006, pp. 2686--2692.

\bibitem{cherubini2009}
A.~Cherubini, M.~Colafrancesco, G.~Oriolo, L.~Freda, and F.~Chaumette,
  ``{Comparing appearance-based controllers for nonholonomic navigation from a
  visual memory},'' in \emph{ICRA 2009 Workshop on safe navigation in open and
  dynamic environments: application to autonomous vehicles}, Kobe, Japan, 2009.

\bibitem{krajnik2010}
T.~Krajn\'ik, J.~Faigl, V.~Von\'sek, K.~Ko\v{s}nar, and M.~K.~L.
  P\v{r}eu\v{c}il, ``{Simple yet stable bearing‐only navigation},''
  \emph{Journal of Field Robotics}, vol.~27, no.~5, pp. 511--533, 2010.

\bibitem{sprunk2013}
C.~Sprunk, G.~D. Tipaldi, A.~Cherubini, and W.~Burgard, ``Lidar-based
  teach-and-repeat of mobile robot trajectories,'' in \emph{In Proc. of
  IEEE/RSJ International Conference on Intelligent Robots and Systems (IROS)},
  2013.

\bibitem{nitsche2014}
M.~Nitsche, T.~Pire, T.~Krajn{\'i}k, M.~Kulich, and M.~Mejail, ``{Monte Carlo
  Localization for Teach-and-Repeat Feature-Based Navigation},'' in
  \emph{{Advances in 15th Annual Conference on Autonomous Robotics Systems}},
  2014, pp. 13--24.

\bibitem{montemerlo2003}
M.~Montemerlo, \emph{{FastSLAM: A Factored Solution to the Simultaneous
  Localization and Mapping Problem With Unknown Data Association}}.\hskip 1em
  plus 0.5em minus 0.4em\relax Carnegie Mellon University: Doctor Thesis, 2003.

\bibitem{baum1966}
L.~E. Baum and T.~Petrie, ``{Statistical Inference for Probabilistic Functions
  of Finite State Markov Chains},'' \emph{The Annals of Mathematical
  Statistics}, vol.~37, no.~6, pp. 1554--1563, 1966.

\bibitem{bishop2006}
C.~M. Bishop, \emph{{Pattern Recognition and Machine Learning}}.\hskip 1em plus
  0.5em minus 0.4em\relax Springer, 2006.

\bibitem{sugiura2011}
K.~Sugiura, N.~Iwahashi, H.~Kashioka, and S.~Nakamura, ``{Learning, Generation
  and Recognition of Motions by Reference-Point-Dependent Probabilistic
  Models},'' \emph{Advanced Robotics}, vol.~25, no. 6-7, pp. 825--848, 2011.

\bibitem{dawood2014}
F.~Dawood and C.~K. Loo, ``{Humanoid Behaviour Learning through Visuomotor
  Association by Self-Imitation},'' in \emph{2014 Joint 7th International
  Conference on Soft Computing and Intelligent Systems (SCIS) and 15th
  International Symposium on Advanced Intelligent Systems (ISIS)}, 2014, pp.
  922--929.

\bibitem{baxter2010}
R.~Baxter, N.~M. Robertson, and D.~Lane, ``Probabilistic behaviour signatures:
  Feature-based behaviour recognition in data-scarce domains,'' in \emph{Proc.
  of 13th International Conference on Information Fusion (Fusion 2010)}, 2010,
  pp. 1--8.

\bibitem{mushtaq2012}
A.~Mushtaq and C.~Hui-Lee, ``{An integrated approach to feature compensation
  combining particle filters and hidden Markov models for robust speech
  recognition},'' in \emph{Proc. of IEEE International Conference on Acoustics,
  Speech and Signal Processing (ICASSP)}, 2012, pp. 4757--4760.

\bibitem{ueda2016ias}
R.~Ueda, K.~Mizuta, H.~Yamakawa, and H.~Okada, ``{Particle Filter on Episode
  for Learning Decision Making Rule},'' in \emph{{Proc. of The 14th
  International Conference on Intelligent Autonomous Systems (IAS-14)}},
  Shanghai, China, 2016.

\bibitem{mayer2006}
H.~Mayer, F.~Gomez, D.~Wierstra, I.~Nagy, A.~Knoll, and J.~Schmidhuber, ``{A
  System for Robotic Heart Surgery that Learns to Tie Knots Using Recurrent
  Neural Networks},'' in \emph{Proceedings of IEEE/RSJ International Conference
  on Intelligent Robots and Systems}, 2006, pp. 543--548.

\bibitem{hochreiter1997}
S.~Hochreiter and J.~Schmidhuber, ``{Long Short-Term Memory},'' \emph{Neural
  Computation}, vol.~9, no.~8, pp. 1735--1780, 1997.

\bibitem{graves2005}
A.~Graves and J.~Schmidhuber, ``{Framewise phoneme classification with
  bidirectional LSTM networks},'' in \emph{Proceedings of IEEE International
  Joint Conference on Neural Networks}, 2005, pp. 2047--2052.

\bibitem{ueda2018ros}
R.~Ueda, \emph{{Learning ROS robot programming with Raspberry Pi}}.\hskip 1em
  plus 0.5em minus 0.4em\relax Nikkei BP, 2018.

\bibitem{kawewong2013}
A.~Kawewong, N.~Tongprasit, and O.~Hasegawa, ``{A speeded-up online incremental
  vision-based loop-closure detection for long-term SLAM},'' \emph{Advanced
  Robotics}, vol.~17, no.~27, pp. 1325--1336, 2013.

\bibitem{saito2018}
A.~Saito and R.~Ueda, ``{Clustering of Memory Sequence for Particle Filter on
  Episode},'' in \emph{Proceedings of the 2018 JSME Conference on Robotics and
  Mechatronics}, Kitakyushu, Japan, 2018 (in Japanese), pp. 1A1--C13.

\end{thebibliography}


\end{document}